\pdfoutput=1

\documentclass[11pt]{article}

\usepackage{acl}

\usepackage{times}
\usepackage{latexsym}

\usepackage[T1]{fontenc}

\usepackage[utf8]{inputenc}

\usepackage{microtype}

\usepackage{graphicx}
\usepackage{tipa}
\usepackage{amssymb}
\usepackage{amsmath}
\usepackage{multirow}
\usepackage{booktabs}
\usepackage{enumitem}

\newcommand{\comment}[1]{}

\newcommand{\ab}[1]{\textcolor{red}{Arnab: #1}}

\title{Automated Cognate Detection as a Supervised Link Prediction Task with Cognate Transformer}

\author{V.S.D.S. Mahesh Akavarapu \and Arnab Bhattacharya \\
  Dept. of Computer Science and Engineering \\
  Indian Institute of Technology Kanpur \\
  \texttt{maheshak@cse.iitk.ac.in, arnabb@cse.iitk.ac.in}
}

\begin{document}

\maketitle

\begin{abstract}
Identification of cognates across related languages is one of the primary problems in historical linguistics. Automated cognate identification is helpful for several downstream tasks including identifying sound correspondences, proto-language reconstruction, phylogenetic classification, etc. Previous state-of-the-art methods for cognate identification are mostly based on distributions of phonemes computed across multilingual wordlists and make little use of the cognacy labels that define links among cognate clusters. In this paper, we present a transformer-based architecture inspired by computational biology for the task of automated cognate detection. Beyond a certain amount of supervision, this method performs better than the existing methods, and shows steady improvement with further increase in supervision, thereby proving the efficacy of utilizing the labeled information. We also demonstrate that accepting multiple sequence alignments as input and having an end-to-end architecture with link prediction head saves much computation time while simultaneously yielding superior performance.
\end{abstract}


\section{Introduction}
\label{sec:intro}

Words in genetically related languages with same descendance from a common ancestral language are termed as \emph{cognates}. For example, Sanskrit \emph{bhava} and English \emph{be} are cognates reconstructed as \emph{*b\textsuperscript{h}ewH-} in ancestral Proto-Indo-European. Within historical linguistics, assembling potential cognates forms an essential step in the comparative method to proceed to further stages such as formulation of sound laws, reconstruction of proto-language, phylogenetic reconstruction, etc. \cite{campbell2013historical}. Cognate identification has been traditionally carried out by tedious manual cross-comparisons of lexica across several concepts or meanings; this often requires sufficient linguistic expertise in the languages that are being compared. \emph{Automated cognate detection} attempts to alleviate manual labor and, thus, assists a historical linguist to quickly produce high-quality etymologies required for downstream tasks.

Over the past decade, several methods for automated cognate detection, mostly using sequence alignment and other techniques inspired by bioinformatics and evolutionary biology \cite{list2017potential}, have appeared. The best-performing methods primarily depend on similarity scores computed from distributions of phonemes in multilingual wordlists \cite{rama-list-2019-automated} and make little or no use of the cognacy labels except for a clustering task at the end. In this paper, we advocate for a supervised learning scenario that utilizes the labeled information to the fullest. We demonstrate that such a scenario combined with the representational power of an appropriate deep neural network architecture can outperform previous methods above a certain amount of supervision. We also demonstrate that such a model is also capable of transfer learning. In other words, once trained on some data, it can perform well on any dataset unseen so far with little additional supervision.

The typical procedure followed by the state-of-the-art methods for this problem is as follows. In each language family, attested words from all languages that have the same meaning, i.e., \emph{concept}, are clustered based on the pairwise similarity measures computed by the respective procedure. We propose a different approach where instead of clustering based on pairwise similarity we directly take input a \emph{multiple sequence alignment (MSA)} of words of the same concept and predict linkage via an end-to-end architecture. This approach proves to be much better in performance and much faster than clustering from independent pairwise similarity measures.

Many of the algorithms in computational historical linguistics are heavily drawn or inspired by computational biology. Continuing the trend, we adopt Cognate Transformer \cite{akavarapu-bhattacharya-2023-cognate}, which yielded state-of-the-art performance in automated phonological reconstruction task, as the base architecture. Cognate Transformer was adapted from MSA Transformer \cite{rao2021msa}, a protein language model that excels in contact predictions. We additionally append to this architecture layers consisting of triangular multiplication and triangular attention modules inspired by Alphafold2 \cite{jumper2021highly}, the state-of-the-art protein structure predictor, where the modules roughly capture triangle inequalities among the distances between amino acid residues. For our task, we applied these modules for capturing transitivity property among linkages in cognate clusters. We find that the addition of this particular module has a significant share in the performance of the overall architecture.

Our key contributions are as follows:

\begin{enumerate}

    \item Firstly, we propose a supervised method for automated cognate detection that outperforms existing methods with sufficient supervision with likely improvement on further supervision, thus utilizing the labeled data much more efficiently than previous models while also demonstrating few-concept (akin to few-shot) learning.

    \item Secondly, our method consists of an end-to-end architecture that avoids independent pairwise computations by accepting MSA as input and directly predicting cluster linkages, which proves to be more efficient in terms of both performance and time than a pairwise approach.

    \item Thirdly, we incorporate into the architecture of Cognate Transformer additional modules to capture transitivity property among cognate cluster linkages which has a positive effect on overall performance.

\end{enumerate}

The rest of the paper is organized as follows. Related work is mentioned in \S\ref{sec:rel}. The problem statement is elaborated in \S\ref{sec:prob}. The methodology is described in \S\ref{sec:meth}. The details of the experimental setup including the datasets used, previous baselines, and evaluation measures are described in \S\ref{sec:exp}. The results of experiments and ablation studies along with error analyses and discussions are given in \S\ref{sec:res}. Finally, the article is concluded in \S\ref{sec:conc}.

\section{Related Work}
\label{sec:rel}

Computational historical linguistics is a young field that emerged over the past two decades. Notable works that lead to significant progress in automatic cognate detection are as follows. Consonant Class Method of \citet{turchin2010analyzing} deems two words as cognate if the first two consonants fall under the same consonant class. In Sound-Class-based phonetic alignment (SCA) of \citet{list2010sca}, pairwise phoneme sequences are aligned and scored for similarity using sound classes that extend consonant classes. LexStat \citep{list-2012-lexstat} aligned and scored pairwise sequences using language phonemic-specific distributions combined with SCA-based scores. The pairwise similarities thus obtained are clustered using UPGMA \cite{Sokal1958ASM}. The previous state-of-the-art results are attributed to LexStat combined with Infomap clustering \cite{list2017potential}. Equivalent performance was also reported in \citet{rama-2018-similarity} using Chinese Restaurant Clustering. An expectation-maximization method over pairwise phonemic distributions is also found to yield similar performance \cite{macsween-caines-2020-expectation}. Information-weighted similarity measure was proposed by \cite{dellert-2018-combining} which reported a slight increase in evaluation scores over LexStat, albeit tested only on one dataset.

Supervised algorithms include the Siamese-CNN-based model by \citet{rama-2016-siamese} which performs binary classification on a given pair of words. \citet{jager-etal-2017-using} employ SVM on top of LexStat and Point-wise Mutual Information (PMI) measures that yield performance similar to that of LexStat-Infomap.

There exist several other works often performing supervised pairwise classification and incorporating multilingual language models such as those of \citet{kanojia-etal-2020-harnessing, kanojia-etal-2021-cognition} and \citet{nath-etal-2022-phonetic}. Despite brilliantly employing pre-trained multilingual language models, these cannot be applied for ancient languages like Ancient Greek, Gothic, etc., or highly low-resource and endangered languages like those of the Americas where one does find wordlists of sufficient size but not enough text to pre-train language models for sake of performing historical linguistic tasks computationally. Another related task is that of cognate and derivate detection \cite{rani-etal-2023-findings}, which is essentially a word-pair classification task. These tasks have a slightly different setup than the problem at hand since the clustering step is not involved.

Cognate Transformer \cite{akavarapu-bhattacharya-2023-cognate} that achieves the best performance on phonological reconstruction tasks employs a transformer-like architecture with row-wise and column-wise attentions to efficiently operate over MSAs. This model was adapted from an evolutionary biological model called MSA Transformer \citep{rao2021msa} which acts on protein sequences. Vanilla Transformer architecture was also used in \citet{kim-etal-2023-transformed} for proto-language reconstruction. Although we employ Cognate Transformer, it should be well noted that the problem we are addressing is that of cognate detection which is quite different from that of proto-language reconstruction. The aforementioned transformer-based models address the latter problem.

\section{Automated Cognate Detection}
\label{sec:prob}

The automated cognate detection problem statement is described here as follows. The gold data for a language family $F$, comprising of related languages $L_1, L_2, \ldots \in F$, consists of words over several concepts, i.e., meanings, say $M_1, M_2, \ldots,$ etc. Each word is a sequence of phonemes. For each concept $M_m$, there are words $W_{i}^{m}$ for several languages $L_i$ in that family, where $W_{i}^{m}$ is a word of a language $L_i$ in concept $M_m$. Words in each concept are associated with labels say $c_i^m \in \mathbb{N}$ which indicate the cluster to which they belong. A single such cluster of words is called a \emph{cognate set}. We also define links $l_{ij}^{k} \in \{0,1\}$ between languages $L_i$ and $L_j$ for a concept $M_m$ which indicate if the corresponding words are cognates i.e., have the same cluster label. In other words,
\begin{align}
l_{ij}^m =
\begin{cases}
1 & \text{if } c_i^m = c_j^m \\
0 & \text{if } c_i^m \neq c_j^m
\end{cases}
\end{align}
The goal of automated cognate detection is to correctly cluster a given set of words that mean a single concept in a language family. In a supervised setting, the aim is to predict the linkages correctly. 

For an illustration of the overall problem, consider the Indo-European language family and the concept of `all'. The attested lexica in the member languages are Sanskrit \emph{s\'{a}rve} (Vedic \emph{v\'{i}\'{s}ve}), Greek (Ancient) \emph{h\'{o}los}, Latin \emph{omnes}, German \emph{alle}, English \emph{all}, Russian \emph{vse}, Czech \emph{vše}, etc. Among these Vedic \emph{v\'{i}\'{s}ve}, Russian \emph{vse}, Czech \emph{vše} form a cluster, i.e., a cognate set while Sanskrit \emph{s\'{a}rve} and Greek \emph{h\'{o}la} form another cognate set. Similarly, English and German word forms form another cognate set. The input data is present in IPA transcription format. Roman transliterated forms are presented here only for demonstration.

\section{Methodology}
\label{sec:meth}

\begin{figure*}[t]
    \includegraphics[width=\textwidth]{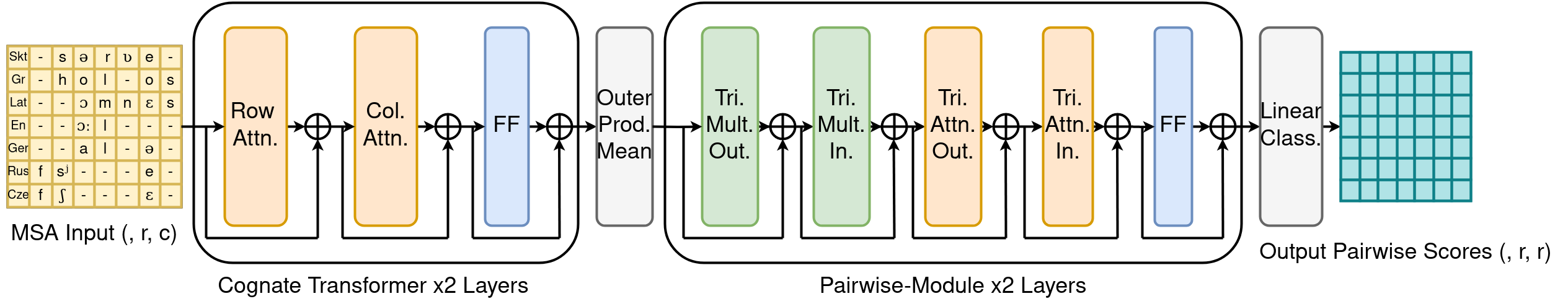}
    \caption{Architecture of Cognate Transformer with Triangle Multiplication and Attention modules}
    \label{fig:arch}
\end{figure*}

The overall workflow is described as follows. Given some words from different languages for a concept in a language family, the words are first aligned (\S\ref{subsec:msa}), then converted into tokens and passed into the cognate transformer (\S\ref{subsec:cogtran}), whose outputs are converted into pairwise (along language axis) representations by outer product mean module (\S\ref{subsec:outprod}), which are then passed into the layers of pairwise module (\S\ref{subsec:pairwise}) whose outputs are classified into two labels 0 or 1 indicating the pairwise linkage among the languages (\S\ref{subsec:class}). Since the linkage information is known in the form of cognacy labels, the architecture described can be thus trained end-to-end. The overall architecture is illustrated in Figure~\ref{fig:arch}.

\subsection{MSA input}
\label{subsec:msa}

\begin{table}[t]
    \centering
    \begin{tabular}{l|ccccccc}
         \toprule
         {Skt.} & - & s & \textipa{@} & r & \textipa{V} & e & - \\
         {Gr.} & - & h & o & l & - & o & s \\
         {Lat.} & - & - & \textipa{O} & m & n & \textipa{E} & s \\
         {En.} & - & - & \textipa{O:} & l & - & - & - \\
         {Ger.} & - & - & \textipa{a} & l & - & \textipa{@} & - \\
         {Rus.} & f & s\textsuperscript{j} & - & - & - & e & - \\
         {Cze.} & f & \textipa{S} & - & - & - & \textipa{E} & - \\
         \bottomrule
    \end{tabular}
    \caption{Example of a Multiple Sequence Alignment (MSA) of phoneme sequences}
    \label{tab:msa}
\end{table}

The input words for a concept are aligned together using the SCA method \citep{list2010sca}, where initial pairwise alignments are carried out by using \citet{needleman1970general} with weights based on sound classes which are further progressively merged guided by a UPGMA \cite{Sokal1958ASM} tree based on pairwise distances. Progressive alignment is a widely used method for multiple sequence alignment which forms the basis of popular programs such as ClustalW \citep{thompson2003multiple}. We use the implementation available in LingPy \citep{list2021lingpy}.

The resultant MSA, present in IPA (see Table~\ref{tab:msa}), is converted into ASJP \citep{brown2008automated} representation, a phonemic representation scheme that compacts IPA symbols resulting in lesser vocabulary size. Note that each token in an MSA need not be a single phoneme. In the SCA method, consecutive vowels are combined into one token. Language information is passed as the initial token in each row following \citet{akavarapu-bhattacharya-2023-cognate}. The resultant tokens are mapped to their respective token numbers and padded according to the batch. Thus, a typical input to Cognate Transformer lies in $\mathbb{N}^{b \times r \times c}$ where $b$ is the batch size, $r$ is the maximum number of rows, i.e., the number of words for that batch, and $c$ is the maximum sequence length in the batch.
From here, we ignore the batch dimension and simply consider the input to lie in $\mathbb{N}^{r \times c}$

\subsection{Cognate Transformer}
\label{subsec:cogtran}

Cognate Transformer \citep{akavarapu-bhattacharya-2023-cognate} handles two-dimensional input employing separate row and column attentions (see Figure~\ref{fig:arch}). The input and output have the same dimensions. In other words,
\begin{align}
\textrm{CogTran}: \mathbb{N}^{r \times c} \rightarrow \mathbb{R}^{r \times c \times d}
\end{align}
where $d$ is the hidden size. The outputs of CogTran are converted into pairwise format by the outer product mean module.

\subsection{Outer Product Mean}
\label{subsec:outprod}

In this module, as the name suggests, the outer product is computed along each column, across all rows, and then the mean of outer products is computed across all columns. The transformation to the dimensions are
\begin{align}
\textrm{OutProdMean}: \mathbb{R}^{r \times c \times d} \rightarrow \mathbb{R}^{r \times r \times d}
\end{align}

The intuition is that the multiplication of a pair of transformed embeddings of two tokens in a single position (i.e., column) of two different words (i.e., rows) should roughly indicate the similarity or distance between the two words in that particular position. The mean operation should produce a mean of such similarities across all positions for a pair of words. Hence, the final matrix would represent a pairwise similarity matrix across the words in an MSA.

This module is identical to the one in AlphaFold2 \citep{jumper2021highly} except that the role of rows and columns is interchanged. In other words, in AlphaFold2, the outputs are pairwise representations of amino-acid-residues (along columns) while in our case the outputs are pairwise representations of words (along rows). 

\subsection{Pairwise Module}
\label{subsec:pairwise}

The pairwise module in AlphaFold2, which consists of triangle multiplication and triangle attention updates via both incoming and outgoing edges, is a differentiable workflow to capture triangle inequalities that the distances between amino acid residues should satisfy \citep{jumper2021highly}. In our case, we demand that the link predictions (see \S\ref{sec:prob} for definition) satisfy the transitivity property which can be translated into the following condition
\begin{align}
l_{ik}^{m} \cdot l_{jk}^{m} = l_{ij}^{m} \textrm{ if } l_{ik}^{m} + l_{jk}^{m} \neq 0
\end{align}
for languages $L_i$, $L_j$ and $L_k$ in a family $F$ for concept $M_m$. The triangle multiplication update follows a similar equation but without constraint and, hence, is apt for the problem at hand. Combining the updates for both incoming ($i \rightarrow j$) and outgoing edges ($j \rightarrow i$) ensures the symmetry required for pairwise similarities. The pairwise module does not alter the dimensions of the input, i.e.,
\begin{align}
\textrm{PairwiseMod}: \mathbb{R}^{r \times r \times d} \rightarrow \mathbb{R}^{r \times r \times d}
\end{align}

In AlphaFold2, this module along with the MSA module is embedded within the Evoformer module. As of now, it is unclear if such embedding would improve the performance. For this problem, we stack the modules as illustrated in Figure~\ref{fig:arch} for the sake of simplicity and easier ablation tests.

\subsection{Classifier and Clustering}
\label{subsec:class}

The outputs of the pairwise module are passed through a linear layer outputting values for two classes $\{0,1\}$ indicating linkage. Hence, the classifier layer's transformation is summarized as:
\begin{align}
\textrm{Classifier}: \mathbb{R}^{r \times r \times d} \rightarrow \mathbb{R}^{r \times r \times 2}
\end{align}

The softmax probabilities of the outputs $p^m_{ij}$ for $P(l_{ij}^m = 1)$ determine the linkage probabilities. During training, the network is trained with cross-entropy loss. During testing, UPGMA is run for each concept $M_m$ with pairwise similarities as $p^m_{ij}$ flat clustered at a threshold of 0.6, which is determined by a small (5\%) held out validation set during training, to obtain the required clusters.

\section{Experimental Setup}
\label{sec:exp}

In this section, the details of the experiments including datasets, implementation, evaluation metrics, baseline models, etc. are described.

\subsection{Datasets}
\label{subsec:data}

\begin{table}[t]
\centering
\resizebox{\columnwidth}{!}{
\begin{tabular}{lrrrr}
\toprule
\textbf{Family} & \textbf{Meanings} & \textbf{Languages} & \textbf{Cognates} & \textbf{Words} \\
\bottomrule
\multicolumn{5}{c}{\textbf{Training data}} \\
\midrule
AN    & 210                  & 20                   & 2864           & 4358           \\
BAI   & 110                  & 9                    & 285            & 1028           \\
CHN   & 140                  & 15                   & 1189           & 2789           \\
IE    & 207                  & 20                   & 1777           & 4393           \\
JAP   & 200                  & 10                   & 460            & 1986           \\
OU    & 110                  & 21                   & 242            & 2055           \\
\midrule
\textbf{Total} & \multicolumn{1}{l}{} & \multicolumn{1}{l}{} & \textbf{6817}  & \textbf{16609} \\
\bottomrule
\multicolumn{5}{c}{\textbf{Test data}} \\
\midrule
BAH   & 200                  & 24                   & 1055           & 4546           \\
CHN   & 180                  & 18                   & 1231           & 3653           \\
HU    & 139                  & 14                   & 855            & 1668           \\
ROM   & 110                  & 43                   & 465            & 4853           \\
TUJ   & 109                  & 5                    & 179            & 513            \\
URA   & 173                  & 7                    & 870            & 1401           \\
AN    & 210                  & 45                   & 3804           & 9267           \\
AA    & 200                  & 58                   & 1872           & 11827          \\
IE    & 208                  & 42                   & 2157           & 9854           \\
PN    & 183                  & 67                   & 6634           & 12691          \\
ST    & 110                  & 64                   & 1402           & 7074           \\
\midrule
\textbf{Total} & \multicolumn{1}{l}{} & \multicolumn{1}{l}{} & \textbf{19136} & \textbf{67347} \\
\bottomrule
\end{tabular}
}
\caption{Details of the datasets as obtained from \citet{rama-list-2019-automated} indicating the number of concepts, languages, cognate sets, and words.}
\label{tab:data}
\end{table}

The dataset for both training and testing along with the train-test split is taken from \citet{rama-list-2019-automated} which was collected from various publicly available sources. It consists of data from various language families, namely, Austro-Asiatic (AA), Austronesian (AN), Bai (BAI), Bahnaric (BAH), Chinese (CHN), Huon (HU), Indo-European (IE), Japanese (JAP), Ob-Ugrian (OU), Pama-Nyungan (PN), Romance (ROM), Sino-Tibetan (ST), Tujia (TUJ), and Uralic (URA). The statistics of the data are provided in Table \ref{tab:data}.

As is evident from the table, the original training size is disproportionately much lesser than the test size. Many language families in tests such as AA, PN, HU, etc. are completely absent in the training set. We also test the model on increased supervision by augmenting the training data with some proportion of test data. In particular, apart from the original train-test split, we also test by including 12.5\% and 50\% additional test concepts, i.e., approximately 20 and 100 additional test concepts respectively per language family. For both the proportions, data is divided into 5 random splits. Hence, the results reported for 12.5\%+ and 50\%+ proportions are five-fold cross-validated.

\subsection{Implementation Details}
\label{subsec:impl}

The architecture we deploy has two Cognate Transformer layers and two layers of pairwise module (see Figure~\ref{fig:arch}). In the Cognate Transformer, the number of attention heads is also 2. The maximum vocabulary size of the tokenizer is set to 768, while the maximum words and sequence length in an MSA are both set to 256. Both hidden size $d$ and intermediate size, wherever there is projection, are 128. This amounts to a network of about a million parameters. The network was trained with a batch size $b$ of 4 and tested with that of 2. Low batch size is due to the limitation of GPU memory (10\,GB in our case) since MSAs combined in both the dimensions and the pairwise representation layers easily blow up the memory. The training was performed using AdamW optimizer \citep{loshchilov2017decoupled} with learning rate 1e-3 as implemented by HuggingFace \citep{wolf-etal-2020-transformers}. During testing, the pairwise softmax probabilities (similarities with 1 being the most similar) are used for flat clustering using UPGMA at a threshold of 0.6, arrived through held-out validation from the train set (5\%). The total time taken for one run of train and test is less than 15 minutes on GPU. This is much smaller when compared to the models that operate on a pair of words at a time instead of on an MSA. The code is made publicly available\footnote{\url{https://github.com/mahesh-ak/CogDetect}}.

\subsection{Evaluation Metrics}
\label{subsec:eval}

The outputs of the entire algorithm are clusters (see \S\ref{subsec:class}), i.e., every word gets a cluster label assigned which is to be compared with the gold cluster labels. The usual F1 score is not a proper measure since the assigned cluster label is not important; rather, members of the same cognate set must get assigned to the same cluster. Hence, the B-Cubed F1 score \citep{amigo2009comparison} is the appropriate evaluation measure; it has been employed in the previous works for this problem as well. We use the implementation available in LingPy \citep{list2021lingpy}.

\subsection{Baseline Models}
\label{subsec:base}

\subsubsection{LexStat-Infomap}

We label the model defined so far as \emph{CogTran2}. The foremost base model with which we compare the performance of CogTran2 is LexStat-Infomap \citep{list2017potential} whose performance is more or less the state-of-the-art as discussed in \S\ref{sec:rel}. The original model employs 10,000 permutations between each language pair in a family to obtain language-specific distributions. Hence, this method requires significant test data to be known beforehand to preprocess. We call this model as \emph{LexStInf10K}. This method takes more than 2 hours on a CPU to obtain results on one test set. Hence we also report for the model that has the number of runs as 1000, which we label as \emph{LexStInf1K} which takes less than 15 CPU minutes. These are imported from LingPy \cite{list2021lingpy}.

\subsubsection{SCA}

We also test on SCA-based model \citep{list2010sca} where a pairwise distance depends on sound classes and alignment. Since it does not depend on any sort of computation such as language-specific distributions, this is the fastest method and, unlike LexStat-Infomap, can be run on any unseen data. We label this as \emph{SCA}. For both LexStat-Informap and SCA, we use the flat cluster thresholds 0.6 and 0.45 respectively, as mentioned in \citet{list2017potential}, since the training data is the same.

\subsubsection{SVM}

We also compare with the SVM-based model \cite{jager-etal-2017-using}, labeled as \emph{SVM}, and the Siamese-CNN-based model \cite{rama-2016-siamese} as these are supervised models. This model uses LexStat score and PMI scores as primary features and, hence, takes a long time to preprocess data, i.e., about 6 hours when each split is processed in parallel on a CPU when the number of permutations runs is 1000 (for LexStat similarity). Since this is a relatively much longer time, we do not increase the number of runs any further. SVM is trained on pairwise binary classification tasks which give pairwise cognacy probabilities for further clustering. We use publicly available code for this model\footnote{\url{https://github.com/evolaemp/svmcc}}.

\subsubsection{Siamese CNN}

From the proposed Siamese CNN architectures \citep{rama-2016-siamese}, we use the model mentioned as charCNN with language features that show good overall performance among the models that are proposed therein. We label this model as \emph{CharCNN}. The network is trained on pairwise supervised binary classification tasks. The pairwise probabilities of the network are used further for clustering (UPGMA). CharCNN is implemented from scratch in PyTorch closely following the TensorFlow code that was made publicly available by the author \footnote{\url{https://github.com/PhyloStar/SiameseConvNet/}}.

\subsubsection{Ablation Models}

We also test on ablations, namely, without pairwise module which we call simply \emph{CogTran}.

We also test by increasing the number of hidden layers to 4 of this same model which we label as \emph{CogTranL4}.

Further, we test on a variant that does not use input MSA but rather only an alignment of a pair of words at a time akin to all other previous models but unlike CogTran2. In this model, pairwise binary classification is performed which gives probability scores for each pair of words in a concept. Further, clustering (UPGMA) is performed using these pairwise scores. To be more specific, the input is an aligned word pair and the resultant output embeddings are summed before the binary classifier, while in Siamese-CNN \cite{rama-2016-siamese}, the absolute differences of embedding pairs are considered before the classifier layer. We note that summing should not be different since the network can always adjust the signs within embeddings themselves. We call this model \emph{CogTranPair}. For these models, the link prediction is not part of the end-to-end architecture, unlike for the model we propose. As a result, the models are run separately on all possible pairs of words in a concept.

\begin{table*}[ht]
\centering
\resizebox{\textwidth}{!}{
\begin{tabular}{ r l ccccccccccc c }
\toprule
\multirow{2}{*}{\textbf{Data+\%}} & \multirow{2}{*}{\textbf{Method}} & \multicolumn{11}{c}{\textbf{Language Families}} & \multirow{2}{*}{\textbf{Mean}} \\
\cmidrule{3-13}
& & \textbf{BAH} & \textbf{CHN} & \textbf{HU} & \textbf{ROM} & \textbf{TUJ} & \textbf{URA} & \textbf{AN} & \textbf{AA} & \textbf{IE} & \textbf{PN} & \textbf{ST} & \\
\midrule
\multirow{6}{*}{\textbf{+0\%}} & SCA & .864 & .793 & .857 & .873 & .894 & .909 & .775 & .760 & .806 & .709 & .561 & .800 \\
& LexStInf10K & .894 & .857 & \textbf{.883} & .910 & \textbf{.899} & \textbf{.913} & .840 & \textbf{.773} & .826 & \textbf{.845} & .592 & \textbf{.839} \\
& LexStInf1K & .894 & .855 & .873 & .912 & .900 & .907 & .839 & .759 & .818 & .820 & \textbf{.595} & .834 \\
& CharCNN & .759 & .837 & .876 & .666 & .845 & .886 & .698 & .722 & .725 & .784 & .473 & .752 \\
& SVM & .865 & .845 & .860 & \textbf{.927} & \textbf{.899} & \textbf{.913} & \textbf{.845} & .734 & .828 & .782 & .593 & .826 \\
& CogTran2 & .854 & \textbf{.864} & .857 & .907 & .893 & .899 & .786 & .756 & \textbf{.845} & .797 & .572 & .821 \\ 
\midrule
& CharCNN & \begin{tabular}[c]{@{}c@{}}.830\\ (.010)\end{tabular} & \begin{tabular}[c]{@{}c@{}}.847\\ (.006)\end{tabular} & \begin{tabular}[c]{@{}c@{}}.873\\ (.010)\end{tabular} & \begin{tabular}[c]{@{}c@{}}.896\\ (.007)\end{tabular} & \begin{tabular}[c]{@{}c@{}}.892\\ (.015)\end{tabular} & \begin{tabular}[c]{@{}c@{}}.895\\ (.006)\end{tabular} & \begin{tabular}[c]{@{}c@{}}.777\\ (.007)\end{tabular} & \begin{tabular}[c]{@{}c@{}}.752\\ (.007)\end{tabular} & \begin{tabular}[c]{@{}c@{}}.825\\ (.008)\end{tabular} & \begin{tabular}[c]{@{}c@{}}.786\\ (.002)\end{tabular} & \begin{tabular}[c]{@{}c@{}}.535\\ (.025)\end{tabular} & \begin{tabular}[c]{@{}c@{}}.810\\ (.002)\end{tabular} \\
\textbf{+12.5\%} & SVM & \begin{tabular}[c]{@{}c@{}}.878\\ (.006)\end{tabular} & \begin{tabular}[c]{@{}c@{}}.836\\ (.006)\end{tabular} & \begin{tabular}[c]{@{}c@{}}.882\\ (.010)\end{tabular} & \textbf{\begin{tabular}[c]{@{}c@{}}.934\\ (.007)\end{tabular}} & \textbf{\begin{tabular}[c]{@{}c@{}}.919\\ (.005)\end{tabular}} & \textbf{\begin{tabular}[c]{@{}c@{}}.914\\ (.006)\end{tabular}} & \textbf{\begin{tabular}[c]{@{}c@{}}.840\\ (.003)\end{tabular}} & \begin{tabular}[c]{@{}c@{}}.767\\ (.012)\end{tabular} & \begin{tabular}[c]{@{}c@{}}.831\\ (.004)\end{tabular} & \begin{tabular}[c]{@{}c@{}}.765\\ (.012)\end{tabular} & \begin{tabular}[c]{@{}c@{}}.582\\ (.012)\end{tabular} & \begin{tabular}[c]{@{}c@{}}.832\\ (.002)\end{tabular} \\
& CogTran2 & \textbf{\begin{tabular}[c]{@{}c@{}}.884\\ (.004)\end{tabular}} & \textbf{\begin{tabular}[c]{@{}c@{}}.867\\ (.005)\end{tabular}} & \textbf{\begin{tabular}[c]{@{}c@{}}.890\\ (.011)\end{tabular}} & \begin{tabular}[c]{@{}c@{}}.907\\ (.013)\end{tabular} & \begin{tabular}[c]{@{}c@{}}.913\\ (.015)\end{tabular} & \begin{tabular}[c]{@{}c@{}}.904\\ (.006)\end{tabular} & \begin{tabular}[c]{@{}c@{}}.810\\ (.005)\end{tabular} & \textbf{\begin{tabular}[c]{@{}c@{}}.813\\ (.003)\end{tabular}} & \textbf{\begin{tabular}[c]{@{}c@{}}.851\\ (.003)\end{tabular}} & \textbf{\begin{tabular}[c]{@{}c@{}}.804\\ (.007)\end{tabular}} & \textbf{\begin{tabular}[c]{@{}c@{}}.607\\ (.020)\end{tabular}} & \textbf{\begin{tabular}[c]{@{}c@{}}.841\\ (.002)\end{tabular}} \\
\midrule
& CharCNN & \begin{tabular}[c]{@{}c@{}}.876\\ (.011)\end{tabular} & \begin{tabular}[c]{@{}c@{}}.854\\ (.007)\end{tabular} & \begin{tabular}[c]{@{}c@{}}.880\\ (.005)\end{tabular} & \begin{tabular}[c]{@{}c@{}}.914\\ (.012)\end{tabular} & \begin{tabular}[c]{@{}c@{}}.899\\ (.018)\end{tabular} & \begin{tabular}[c]{@{}c@{}}.904\\ (.012)\end{tabular} & \begin{tabular}[c]{@{}c@{}}.795\\ (.008)\end{tabular} & \begin{tabular}[c]{@{}c@{}}.784\\ (.005)\end{tabular} & \begin{tabular}[c]{@{}c@{}}.840\\ (.006)\end{tabular} & \begin{tabular}[c]{@{}c@{}}.785\\ (.011)\end{tabular} & \begin{tabular}[c]{@{}c@{}}.563\\ (.011)\end{tabular} & \begin{tabular}[c]{@{}c@{}}.827\\ (.003)\end{tabular} \\
\textbf{+50\%} & SVM & \begin{tabular}[c]{@{}c@{}}.881\\ (.010)\end{tabular} & \begin{tabular}[c]{@{}c@{}}.838\\ (.009)\end{tabular} & \begin{tabular}[c]{@{}c@{}}.889\\ (.014)\end{tabular} & \textbf{\begin{tabular}[c]{@{}c@{}}.935\\ (.010)\end{tabular}} & \textbf{\begin{tabular}[c]{@{}c@{}}.927\\ (.012)\end{tabular}} & \textbf{\begin{tabular}[c]{@{}c@{}}.914\\ (.009)\end{tabular}} & \textbf{\begin{tabular}[c]{@{}c@{}}.840\\ (.010)\end{tabular}} & \begin{tabular}[c]{@{}c@{}}.779\\ (.009)\end{tabular} & \begin{tabular}[c]{@{}c@{}}.828\\ (.007)\end{tabular} & \begin{tabular}[c]{@{}c@{}}.775\\ (.009)\end{tabular} & \begin{tabular}[c]{@{}c@{}}.577\\ (.019)\end{tabular} & \begin{tabular}[c]{@{}c@{}}.835\\ (.002)\end{tabular} \\
& CogTran2 & \textbf{\begin{tabular}[c]{@{}c@{}}.893\\ (.011)\end{tabular}} & \textbf{\begin{tabular}[c]{@{}c@{}}.878\\ (.005)\end{tabular}} & \textbf{\begin{tabular}[c]{@{}c@{}}.901\\ (.006)\end{tabular}} & \begin{tabular}[c]{@{}c@{}}.921\\ (.015)\end{tabular} & \begin{tabular}[c]{@{}c@{}}.916\\ (.009)\end{tabular} & \textbf{\begin{tabular}[c]{@{}c@{}}.914\\ (.007)\end{tabular}} & \begin{tabular}[c]{@{}c@{}}.823\\ (.006)\end{tabular} & \textbf{\begin{tabular}[c]{@{}c@{}}.832\\ (.008)\end{tabular}} & \textbf{\begin{tabular}[c]{@{}c@{}}.853\\ (.004)\end{tabular}} & \textbf{\begin{tabular}[c]{@{}c@{}}.812\\ (.006)\end{tabular}} & \textbf{\begin{tabular}[c]{@{}c@{}}.644\\ (.015)\end{tabular}} & \textbf{\begin{tabular}[c]{@{}c@{}}.853\\ (.002)\end{tabular}} \\
\bottomrule
\end{tabular}
}
\caption{Results (B-Cubed F-scores) with language families indicated across columns along with standard deviations in parentheses for cross-validated values. The best scores within a specific train-test split are shown in bold.}
\label{tab:res}
\end{table*}

\section{Results}
\label{sec:res}

The results are summarized in Table~\ref{tab:res}. The first column indicates the additional proportion of concepts that is moved from test data to training data. Thus, it roughly indicates the amount of increased supervision. The second column indicates the various methods discussed in \S\ref{subsec:base} compared against the proposed model, \emph{CogTran2}. The rest of the columns indicate the B-Cubed F scores (see \S\ref{subsec:eval}) for various datasets discussed in \S\ref{subsec:data}. The last column indicates the mean B-Cubed F-scores averaged across the aforementioned datasets.

For the additional proportions +12.5\% and +50\%, the reported scores are means along with standard deviations (in parentheses) over the five validation sets (see \S\ref{subsec:data}). Note that the standard deviation for the overall averaged B-Cubed F score is considerably much less than those of individual datasets. This happens since in every run on a train-test split the model may perform high on one dataset or low on the other, yet when it comes to the mean performance it is quite stable.

\comment{
\begin{figure}[t]
    \centering
    \includegraphics[width=1.05\columnwidth]{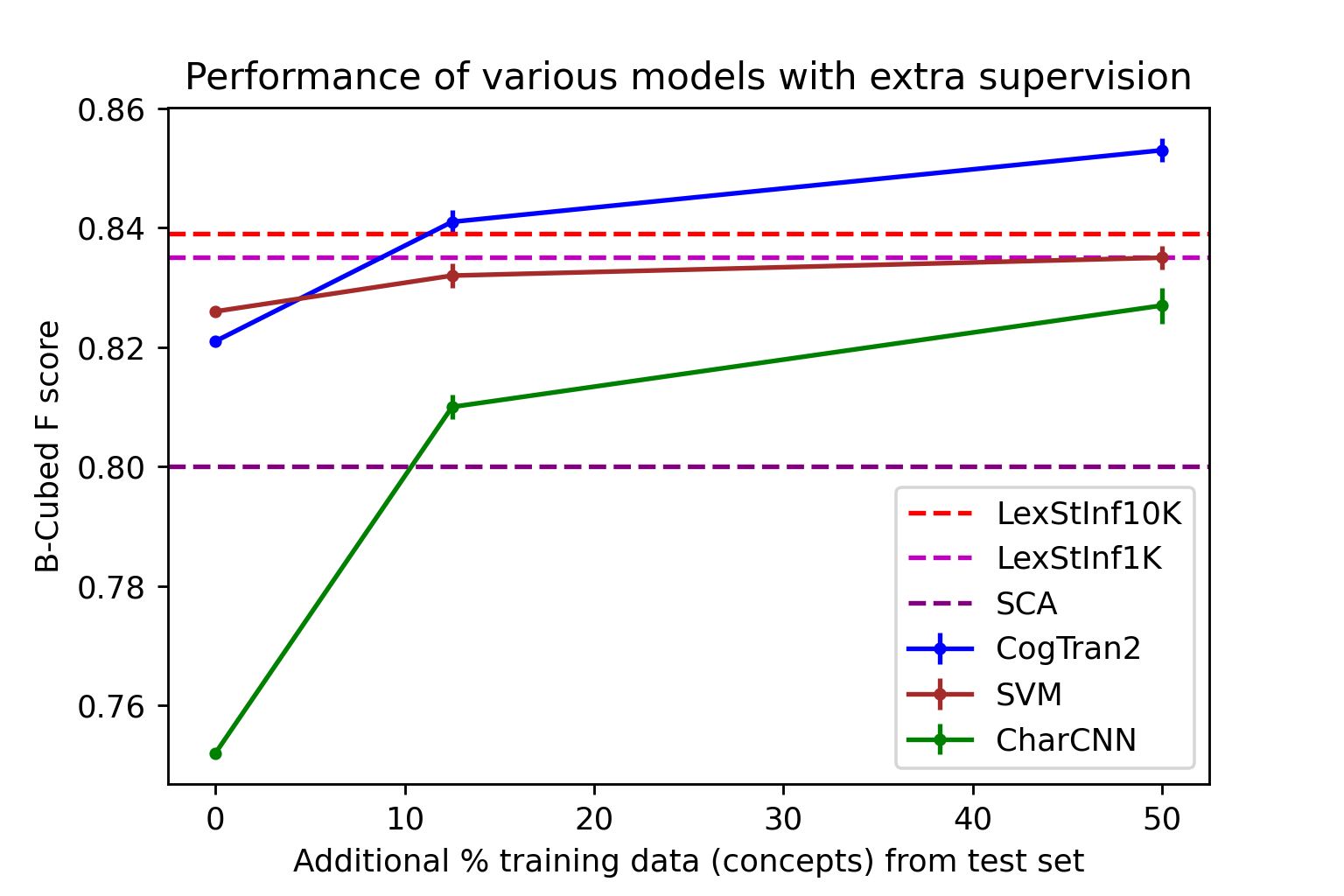}
    \caption{Plot of average B-Cubed F scores}
    \label{fig:bcf}
\end{figure}

We also plot the B-cubed F-scores of these models in Figure~\ref{fig:bcf}. \ab{if the figure conveys the same information as the table, then it is not required.}
}

\subsection{Discussion}

From the results, it is visible that with increased supervision, CogTran2 improves consistently when compared to other supervised methods. At the same time, CogTran2 crosses the previous best LexStInf10K with additional +12.5\% supervision, i.e., with only 20 concepts per family. Since the results of proportions +12.5\% and +50\% are cross-validated, it is possible to compare the performances throughout. Note that LexStat is not a supervised method and, hence, additional supervision does not make sense with it. With zero additional data, CogTran2 surpasses all the other methods on CHN and IE language families since they are present in training as well. While AN data is also present in both sets i.e., train and test, the individual languages do not overlap much as in the case of CHN and IE.

Although SVM beats CogTran2 on +0\% additional data, which is not surprising since this is primarily dependent on LexStInf1K scores, it shows only a little increase in scores with an increase in additional training. Hence, overall, it is behind CogTran2 for the other two proportions. The maximum score of SVM does not appear to be significantly different from its base model LexStInf1K on whose scores it is dependent. We performed student t-tests vis-\`{a}-vis SVM and CogTran2 scores for proportions +12.5\% and +50\%. On whatever dataset CogTran2 leads ahead of SVM, it is statistically significant for a 5\% level of significance, i.e., $p < 0.05$. SVM leads ahead of CogTran2 significantly only on two datasets, namely, Austronesian (AN) and Romance (ROM) in both proportions. The reason for this is unclear as of now. Analysis with linguistic expertise in these languages could possibly unveil the cause.

CharCNN has the disadvantage of not using aligned input. Hence, it lags behind other models as expected (except SCA at extra supervision) despite showing a significant improvement over the additional training data.

Hence, it can be concluded that CogTran2 is the best performing model when there is sufficient labeled data.
It is also likely to show improvement when there is plenty of labeled data. Further, given the availability of GPU and considering the present implementations, CogTran2 is much faster since it starts from MSA and not from independent pairwise computations.

\subsection{Ablation Tests}
\label{subsec:ablres}

\begin{table}[t]
\centering
\resizebox{\columnwidth}{!}{
\begin{tabular}{lccc}
\toprule
\multirow{2}{*}{\textbf{Method}}  & \multicolumn{3}{c}{\textbf{Data Split}} \\
\cmidrule{2-4} 
& \multicolumn{1}{c}{\textbf{+0\%}} & \multicolumn{1}{c}{\textbf{+12.5\%}} & \textbf{+50\%} \\
\midrule
CogTran2 & \multicolumn{1}{c}{\textbf{0.821}} & \multicolumn{1}{c}{\textbf{0.841 (± 0.002)}} & \textbf{0.853 (± 0.002)} \\
CogTran & \multicolumn{1}{c}{0.815} & \multicolumn{1}{c}{0.830 (± 0.002)} & 0.841 (± 0.002) \\
CogTranL4 & \multicolumn{1}{c}{0.806} & \multicolumn{1}{c}{0.830 (± 0.002)} & 0.842 (± 0.004) \\
CogTranPair & \multicolumn{1}{c}{0.779} & \multicolumn{1}{c}{0.813 (± 0.003)} & 0.833 (± 0.001) \\
\bottomrule
\end{tabular}
}
\caption{Mean B-Cubed F scores on various data splits for various ablation models. Standard deviations are indicated in parentheses for the data splits where cross-validation was performed.}
\label{tab:abl}
\end{table}

The results of the ablation tests described in \S\ref{subsec:base} on the data proportions +0\%, +12.5\% and +50\% are presented in Table \ref{tab:abl}. The first column indicates the method and the second column lists the respective B-Cubed F-score averaged over all the datasets. These are mean scores along with standard deviations across all five cross-validated sets. CogTran, which lacks a Pairwise module (\S\ref{subsec:pairwise}), underperforms significantly than CogTran2, which is the model proposed. Also, increasing the number of layers to 4 in CogTranL4 does not help either. Hence, it can be concluded that the Pairwise module alone contributes to further increasing the performance in CogTran2. Further since CogTrainPair, unlike the other two, starts from aligned word pairs akin to all other previous models, and takes input from an aligned word pair and outputs cognacy probability for that pair. Hence, the Pairwise module cannot be incorporated into this setup. 

It is visible that CogTran, which acts on an MSA input performs way better than CogTranPair which acts on aligned word pairs. At the same time, CogTran (< 20 GPU min per split) is much faster than CogTranPair (about 1 GPU hr per split) for the same reason. In other words, let input MSA have $r$ rows and $c$ columns, then CogTranPair acts on all possible pairs of rows hence, in $\mbox{O}{(r^2)}$ steps. On the other hand, CogTran for a single MSA acts only once which results in the speed-up.

\subsection{Error Analysis}
\label{subsec:err}

To understand the working of CogTran2, we attempt to study some of the cluster predictions as follows. For this purpose, we consider CogTran2 trained on +12.5\% proportion and the results on IE (Indo-European) dataset. 

\subsubsection{Sound Correspondences}

The fundamental aspect for comparing two languages is to identify regular sound correspondences \citep{campbell2013historical}. Methods like LexStat \cite{list-2012-lexstat} have built similarity metrics for cognacy judgement between two words giving weightage to both the recurrent sound correspondences as well as phonetic information. In this regard, we note that CogTran2 appears to have learned some recurrent sound correspondences by observing the initial consonant. For example, Proto-Indo-European \textit{*s-} undergoes lenition in Hellenic branch and appears as \textit{h-} is Ancient Greek \citep{mallory2006oxford}. In the dataset we have used, two words occur as instances for this sound change, namely, /\textipa{"hE:lios}/ `sun' and /\textipa{"hals}/ `salt'. Both these words are clustered correctly with their cognates in other daughter languages such as Old Norse /\textipa{so:l}/, Oriya /\textipa{surdZO}/ in case of the concept `sun' and English /\textipa{sO:lt}/, French /\textipa{sEl}/ in case of the concept `salt'. Thus, one may assume that the sound change PIE \textit{*s} > Ancient Greek \textit{h} has been learned by the model. 

Another set of sound changes where position of articulation changes is Grimm's law where Proto-Indo-European hard consonants undergo a chain shift in Germanic family \citep{mallory2006oxford}. For instance, in the velar shift defined by Grimm's law i.e., \textit{*g\textsuperscript{h} > *g > *k > *h} , change in the place of articulation occurs in the sound change \textit{*k > *h}. The model also learns this sound change as supported by the instances mentioned as follows. For the concept `dog', German /\textipa{hUnt}/ has been correctly clustered together with Ancient Greek /\textipa{"kyOn}/ and Old Irish /\textipa{ku:}/. Further, for the concept 'horn', German /\textipa{hOrn}/ and Ancient Greek /keras/ are similarly clustered together correctly. This sound change has been learned by the model to an extent that unrelated German /\textipa{hIm\s{l}}/ and Latin /\textipa{ka\textsubarch{e}lUm}/ meaning `sky' have been classified as cognates. Both the sound changes mentioned above have two instances as examples in the dataset.

On the other hand, Marathi /\textipa{dzaN}/ and Ossetic /\textipa{zon}/ for the concept `know' have been incorrectly classified as different. This happens to be the only example where the phonemes /\textipa{dz}/ and /\textipa{z}/, which fall in different sound classes, co-occur in the respective languages. Hence, it may be concluded that at least two examples are needed to learn a sound change. However, it is desirable to perform a thorough quantitative analysis of recurrent sound changes to support these findings. It could not be performed due to a lack of readily available annotated data for the same.

\subsubsection{Partial Cognacy}

Further, the network seems to consider the entire word and not just the important root in some cases. For example, for the meaning `woman', Old Norse /kve\textipa{n: maDr}/ and Icelandic /k\textsuperscript{h}v\textipa{En ma:Dr}/ have been assigned a different cluster than that of Old Swedish \textipa{/kvin:a/} and Danish /g\textsuperscript{h}\textipa{ven@}/. This is conceivable since affixes cannot be learned to be ignored easily. Detection of sub-word cognates in presence of such affixes is part of \emph{partial cognacy} problem which was dealt in \citet{list-etal-2016-using}. It is, thus, clear that CogTran2, at its present training level, cannot distinguish partial cognates.

\subsubsection{Other Errors}

Many errors are, however, somewhat incomprehensible. For example, in the case of `tooth', Greek /\textipa{"Dondi}/ has been clustered together with English /\textipa{tu:T}/ but not with Italian /\textipa{dEntE}/. There could be a role of root vowel in this particular example. Nevertheless, it is important to understand the source of errors which demands linguistic expertise to identify the bottlenecks of the current models and to improve beyond them.

\section{Conclusions}
\label{sec:conc}

In this paper, we have proposed a Transformer-based model inspired by evolutionary biology for the task of automatic cognate detection. The model can harness efficiently the labeled data and consequently, with sufficient data, outperforms existing approaches that do not make efficient use of supervision data. In particular, better results are obtained with only 20 concepts per family on some of the datasets. To the best of our knowledge, we proposed for the first time in this particular problem a method that directly outputs link probabilities, i.e., pairwise similarities from an input MSA in an end-to-end fashion, unlike all the previous methods which act on aligned pairs of words. We demonstrated through the primary results and ablation studies that this approach of inputting MSA rather than paired alignments results not just in a significant increase in performance but also in drastically reducing the computation time. We have also demonstrated by observing few outputs that the model is capable of learning regular sound changes from just two example instances in the data for a particular sound change.

Evaluation of Cognate Transformer on phylogenetic reconstruction task \citep{rama-etal-2018-automatic} is an unexplored problem and, thus, can be a potential topic of future work.


\section*{Limitations}
\label{sec:lim}

As mentioned in \S\ref{sec:res}, the proposed model lags on the datasets Romance and Austronesian somewhat behind SVM and LexStat-Infomap and on Pama-Nyungan concerning Lexstat-Infomap despite increasing the supervision. While the performance on the Romance dataset is near saturated (>92\%) in any case, the lag in performance on Austronesian and Pama-Nyungan data is an issue that is required to be studied with domain linguistic expertise to understand the bottleneck of this model. Similarly, although our model improves drastically on Sino-Tibetan by 5\% when compared to the previous best, it is an underperforming dataset since the B-Cubed F-scores on all other datasets except this are more than 80\%. Thus, a similar study with linguistic expertise is required to identify the bottleneck of the overall methodologies. Additionally, as mentioned in \S\ref{subsec:impl}, a GPU memory of 10GB could only accommodate a batch of size 4 during training with maximum MSAs, i.e., when the number of languages in a family was 136. Thus, larger GPU storage is required for larger mass comparisons involving more languages under comparisons. As mentioned in \S\ref{subsec:err}, the ability of the model to learn regular sound correspondences has only been determined by anecdotal instances. A more thorough quantitative study is desirable, which requires annotated data for the same. The model also does not account for partial cognacy, i.e., identifying distinctions between exact cognates versus morphologically modified or compounded cognates (see \S\ref{subsec:err}) as addressed in \citet{list-etal-2016-using}. Further, the model is also not tuned at this point to distinguish between true cognates and borrowals.

\section*{Ethics Statement}
\label{sec:ethics}

All the data and code of baseline models used in this paper are obtained from publicly available sources. Hence, we do not see any ethical concerns or conflicts of interest.


\bibliography{anthology,custom}

\begin{thebibliography}{30}
\expandafter\ifx\csname natexlab\endcsname\relax\def\natexlab#1{#1}\fi

\bibitem[{Akavarapu and
  Bhattacharya(2023)}]{akavarapu-bhattacharya-2023-cognate}
V.S.D.S.Mahesh Akavarapu and Arnab Bhattacharya. 2023.
\newblock \href {https://doi.org/10.18653/v1/2023.emnlp-main.423} {Cognate
  transformer for automated phonological reconstruction and cognate reflex
  prediction}.
\newblock In \emph{Proceedings of the 2023 Conference on Empirical Methods in
  Natural Language Processing}, pages 6852--6862, Singapore. Association for
  Computational Linguistics.

\bibitem[{Amig{\'o} et~al.(2009)Amig{\'o}, Gonzalo, Artiles, and
  Verdejo}]{amigo2009comparison}
Enrique Amig{\'o}, Julio Gonzalo, Javier Artiles, and Felisa Verdejo. 2009.
\newblock A comparison of extrinsic clustering evaluation metrics based on
  formal constraints.
\newblock \emph{Information retrieval}, 12:461--486.

\bibitem[{Brown et~al.(2008)Brown, Holman, Wichmann, and
  Velupillai}]{brown2008automated}
Cecil~H Brown, Eric~W Holman, S{\o}ren Wichmann, and Viveka Velupillai. 2008.
\newblock Automated classification of the world's languages: a description of
  the method and preliminary results.
\newblock \emph{Language Typology and Universals}, 61(4):285--308.

\bibitem[{Campbell(2013)}]{campbell2013historical}
Lyle Campbell. 2013.
\newblock \emph{Historical linguistics}.
\newblock Edinburgh University Press.

\bibitem[{Dellert(2018)}]{dellert-2018-combining}
Johannes Dellert. 2018.
\newblock \href {https://aclanthology.org/C18-1264} {Combining
  information-weighted sequence alignment and sound correspondence models for
  improved cognate detection}.
\newblock In \emph{Proceedings of the 27th International Conference on
  Computational Linguistics}, pages 3123--3133, Santa Fe, New Mexico, USA.
  Association for Computational Linguistics.

\bibitem[{J{\"a}ger et~al.(2017)J{\"a}ger, List, and
  Sofroniev}]{jager-etal-2017-using}
Gerhard J{\"a}ger, Johann-Mattis List, and Pavel Sofroniev. 2017.
\newblock \href {https://aclanthology.org/E17-1113} {Using support vector
  machines and state-of-the-art algorithms for phonetic alignment to identify
  cognates in multi-lingual wordlists}.
\newblock In \emph{Proceedings of the 15th Conference of the {E}uropean Chapter
  of the Association for Computational Linguistics: Volume 1, Long Papers},
  pages 1205--1216, Valencia, Spain. Association for Computational Linguistics.

\bibitem[{Jumper et~al.(2021)Jumper, Evans, Pritzel, Green, Figurnov,
  Ronneberger, Tunyasuvunakool, Bates, {\v{Z}}{\'\i}dek, Potapenko
  et~al.}]{jumper2021highly}
John Jumper, Richard Evans, Alexander Pritzel, Tim Green, Michael Figurnov,
  Olaf Ronneberger, Kathryn Tunyasuvunakool, Russ Bates, Augustin
  {\v{Z}}{\'\i}dek, Anna Potapenko, et~al. 2021.
\newblock Highly accurate protein structure prediction with alphafold.
\newblock \emph{Nature}, 596(7873):583--589.

\bibitem[{Kanojia et~al.(2020)Kanojia, Dabre, Dewangan, Bhattacharyya, Haffari,
  and Kulkarni}]{kanojia-etal-2020-harnessing}
Diptesh Kanojia, Raj Dabre, Shubham Dewangan, Pushpak Bhattacharyya, Gholamreza
  Haffari, and Malhar Kulkarni. 2020.
\newblock \href {https://doi.org/10.18653/v1/2020.coling-main.119} {Harnessing
  cross-lingual features to improve cognate detection for low-resource
  languages}.
\newblock In \emph{Proceedings of the 28th International Conference on
  Computational Linguistics}, pages 1384--1395, Barcelona, Spain (Online).
  International Committee on Computational Linguistics.

\bibitem[{Kanojia et~al.(2021)Kanojia, Sharma, Ghodekar, Bhattacharyya,
  Haffari, and Kulkarni}]{kanojia-etal-2021-cognition}
Diptesh Kanojia, Prashant Sharma, Sayali Ghodekar, Pushpak Bhattacharyya,
  Gholamreza Haffari, and Malhar Kulkarni. 2021.
\newblock \href {https://doi.org/10.18653/v1/2021.eacl-main.288}
  {Cognition-aware cognate detection}.
\newblock In \emph{Proceedings of the 16th Conference of the European Chapter
  of the Association for Computational Linguistics: Main Volume}, pages
  3281--3292, Online. Association for Computational Linguistics.

\bibitem[{Kim et~al.(2023)Kim, Chang, Cui, and
  Mortensen}]{kim-etal-2023-transformed}
Young~Min Kim, Kalvin Chang, Chenxuan Cui, and David~R. Mortensen. 2023.
\newblock \href {https://doi.org/10.18653/v1/2023.acl-short.3} {Transformed
  protoform reconstruction}.
\newblock In \emph{Proceedings of the 61st Annual Meeting of the Association
  for Computational Linguistics (Volume 2: Short Papers)}, pages 24--38,
  Toronto, Canada. Association for Computational Linguistics.

\bibitem[{List(2010)}]{list2010sca}
Johann-Mattis List. 2010.
\newblock Sca: Phonetic alignment based on sound classes.
\newblock In \emph{European Summer School in Logic, Language and Information},
  pages 32--51. Springer.

\bibitem[{List(2012)}]{list-2012-lexstat}
Johann-Mattis List. 2012.
\newblock \href {https://aclanthology.org/W12-0216} {{L}ex{S}tat: Automatic
  detection of cognates in multilingual wordlists}.
\newblock In \emph{Proceedings of the {EACL} 2012 Joint Workshop of {LINGVIS}
  {\&} {UNCLH}}, pages 117--125, Avignon, France. Association for Computational
  Linguistics.

\bibitem[{List and Forkel(2021)}]{list2021lingpy}
Johann-Mattis List and Robert Forkel. 2021.
\newblock \href {https://lingpy.org} {Lingpy. a python library for historical
  linguistics. version 2.6.9}.

\bibitem[{List et~al.(2017)List, Greenhill, and Gray}]{list2017potential}
Johann-Mattis List, Simon~J Greenhill, and Russell~D Gray. 2017.
\newblock The potential of automatic word comparison for historical
  linguistics.
\newblock \emph{PloS one}, 12(1):e0170046.

\bibitem[{List et~al.(2016)List, Lopez, and Bapteste}]{list-etal-2016-using}
Johann-Mattis List, Philippe Lopez, and Eric Bapteste. 2016.
\newblock \href {https://doi.org/10.18653/v1/P16-2097} {Using sequence
  similarity networks to identify partial cognates in multilingual wordlists}.
\newblock In \emph{Proceedings of the 54th Annual Meeting of the Association
  for Computational Linguistics (Volume 2: Short Papers)}, pages 599--605,
  Berlin, Germany. Association for Computational Linguistics.

\bibitem[{Loshchilov and Hutter(2017)}]{loshchilov2017decoupled}
Ilya Loshchilov and Frank Hutter. 2017.
\newblock Decoupled weight decay regularization.
\newblock \emph{arXiv preprint arXiv:1711.05101}.

\bibitem[{MacSween and Caines(2020)}]{macsween-caines-2020-expectation}
Roddy MacSween and Andrew Caines. 2020.
\newblock \href {https://doi.org/10.18653/v1/2020.conll-1.38} {An expectation
  maximisation algorithm for automated cognate detection}.
\newblock In \emph{Proceedings of the 24th Conference on Computational Natural
  Language Learning}, pages 476--485, Online. Association for Computational
  Linguistics.

\bibitem[{Mallory and Adams(2006)}]{mallory2006oxford}
James~P Mallory and Douglas~Q Adams. 2006.
\newblock \emph{The Oxford introduction to proto-Indo-European and the
  proto-Indo-European world}.
\newblock Oxford University Press, USA.

\bibitem[{Nath et~al.(2022)Nath, Ghosh, and
  Krishnaswamy}]{nath-etal-2022-phonetic}
Abhijnan Nath, Rahul Ghosh, and Nikhil Krishnaswamy. 2022.
\newblock \href {https://aclanthology.org/2022.vardial-1.5} {Phonetic,
  semantic, and articulatory features in {A}ssamese-{B}engali cognate
  detection}.
\newblock In \emph{Proceedings of the Ninth Workshop on NLP for Similar
  Languages, Varieties and Dialects}, pages 41--53, Gyeongju, Republic of
  Korea. Association for Computational Linguistics.

\bibitem[{Needleman and Wunsch(1970)}]{needleman1970general}
Saul~B Needleman and Christian~D Wunsch. 1970.
\newblock A general method applicable to the search for similarities in the
  amino acid sequence of two proteins.
\newblock \emph{Journal of molecular biology}, 48(3):443--453.

\bibitem[{Rama(2016)}]{rama-2016-siamese}
Taraka Rama. 2016.
\newblock \href {https://aclanthology.org/C16-1097} {{S}iamese convolutional
  networks for cognate identification}.
\newblock In \emph{Proceedings of {COLING} 2016, the 26th International
  Conference on Computational Linguistics: Technical Papers}, pages 1018--1027,
  Osaka, Japan. The COLING 2016 Organizing Committee.

\bibitem[{Rama(2018)}]{rama-2018-similarity}
Taraka Rama. 2018.
\newblock \href {https://doi.org/10.18653/v1/K18-1027} {Similarity dependent
  {C}hinese restaurant process for cognate identification in multilingual
  wordlists}.
\newblock In \emph{Proceedings of the 22nd Conference on Computational Natural
  Language Learning}, pages 271--281, Brussels, Belgium. Association for
  Computational Linguistics.

\bibitem[{Rama and List(2019)}]{rama-list-2019-automated}
Taraka Rama and Johann-Mattis List. 2019.
\newblock \href {https://doi.org/10.18653/v1/P19-1627} {An automated framework
  for fast cognate detection and {B}ayesian phylogenetic inference in
  computational historical linguistics}.
\newblock In \emph{Proceedings of the 57th Annual Meeting of the Association
  for Computational Linguistics}, pages 6225--6235, Florence, Italy.
  Association for Computational Linguistics.

\bibitem[{Rama et~al.(2018)Rama, List, Wahle, and
  J{\"a}ger}]{rama-etal-2018-automatic}
Taraka Rama, Johann-Mattis List, Johannes Wahle, and Gerhard J{\"a}ger. 2018.
\newblock \href {https://doi.org/10.18653/v1/N18-2063} {Are automatic methods
  for cognate detection good enough for phylogenetic reconstruction in
  historical linguistics?}
\newblock In \emph{Proceedings of the 2018 Conference of the North {A}merican
  Chapter of the Association for Computational Linguistics: Human Language
  Technologies, Volume 2 (Short Papers)}, pages 393--400, New Orleans,
  Louisiana. Association for Computational Linguistics.

\bibitem[{Rani et~al.(2023)Rani, Goswami, Doyle, Fransen, Stearns, and
  McCrae}]{rani-etal-2023-findings}
Priya Rani, Koustava Goswami, Adrian Doyle, Theodorus Fransen, Bernardo
  Stearns, and John~P. McCrae. 2023.
\newblock \href {https://doi.org/10.18653/v1/2023.sigtyp-1.13} {Findings of the
  {SIGTYP} 2023 shared task on cognate and derivative detection for
  low-resourced languages}.
\newblock In \emph{Proceedings of the 5th Workshop on Research in Computational
  Linguistic Typology and Multilingual NLP}, pages 126--131, Dubrovnik,
  Croatia. Association for Computational Linguistics.

\bibitem[{Rao et~al.(2021)Rao, Liu, Verkuil, Meier, Canny, Abbeel, Sercu, and
  Rives}]{rao2021msa}
Roshan~M Rao, Jason Liu, Robert Verkuil, Joshua Meier, John Canny, Pieter
  Abbeel, Tom Sercu, and Alexander Rives. 2021.
\newblock Msa transformer.
\newblock In \emph{International Conference on Machine Learning}, pages
  8844--8856. PMLR.

\bibitem[{Sokal and Michener(1958)}]{Sokal1958ASM}
Robert~R. Sokal and Charles~Duncan Michener. 1958.
\newblock A statistical method for evaluating systematic relationships.
\newblock \emph{University of Kansas science bulletin}, 38:1409--1438.

\bibitem[{Thompson et~al.(2003)Thompson, Gibson, and
  Higgins}]{thompson2003multiple}
Julie~D Thompson, Toby~J Gibson, and Des~G Higgins. 2003.
\newblock Multiple sequence alignment using clustalw and clustalx.
\newblock \emph{Current protocols in bioinformatics}, (1):2--3.

\bibitem[{Turchin et~al.(2010)Turchin, Peiros, and
  Gell-Mann}]{turchin2010analyzing}
Peter Turchin, Ilia Peiros, and Murray Gell-Mann. 2010.
\newblock Analyzing genetic connections between languages by matching consonant
  classes.
\newblock \emph{Journal of Language Relationship}, (5 (48)):117--126.

\bibitem[{Wolf et~al.(2020)Wolf, Debut, Sanh, Chaumond, Delangue, Moi, Cistac,
  Rault, Louf, Funtowicz, Davison, Shleifer, von Platen, Ma, Jernite, Plu, Xu,
  Le~Scao, Gugger, Drame, Lhoest, and Rush}]{wolf-etal-2020-transformers}
Thomas Wolf, Lysandre Debut, Victor Sanh, Julien Chaumond, Clement Delangue,
  Anthony Moi, Pierric Cistac, Tim Rault, Remi Louf, Morgan Funtowicz, Joe
  Davison, Sam Shleifer, Patrick von Platen, Clara Ma, Yacine Jernite, Julien
  Plu, Canwen Xu, Teven Le~Scao, Sylvain Gugger, Mariama Drame, Quentin Lhoest,
  and Alexander Rush. 2020.
\newblock \href {https://doi.org/10.18653/v1/2020.emnlp-demos.6} {Transformers:
  State-of-the-art natural language processing}.
\newblock In \emph{Proceedings of the 2020 Conference on Empirical Methods in
  Natural Language Processing: System Demonstrations}, pages 38--45, Online.
  Association for Computational Linguistics.

\end{thebibliography}


\end{document}